\documentclass{article}

\PassOptionsToPackage{round,sort}{natbib}  %

 \usepackage[main, final]{neurips_2026}

\usepackage[utf8]{inputenc} %
\usepackage[T1]{fontenc}    %
\usepackage{hyperref}       %
\usepackage{url}            %
\usepackage{booktabs}       %
\usepackage{amsmath}        %
\usepackage{amsfonts}       %
\usepackage{amssymb}        %
\usepackage{nicefrac}       %
\usepackage{microtype}      %
\usepackage[dvipsnames]{xcolor}  %
\usepackage{graphicx}       %
\usepackage{float}          %
\usepackage{tikz}           %
\usetikzlibrary{positioning, arrows.meta, fit, calc, shapes.geometric}

\hypersetup{
  colorlinks=true,
  linkcolor=blue,        %
  citecolor=blue,        %
  urlcolor=OliveGreen,   %
  filecolor=OliveGreen,
}

\newcommand{\Deff}{D_{\mathrm{eff}}}

\newcommand{\Hsol}{H^{\star}}
\newcommand{\dePo}{\textsc{dePo}}

\title{Inside the Loop: A Mechanistic Study of Looped Transformers on Permutation Traversal}

\newif\ifarxiv
\arxivtrue
\ifarxiv
\author{
  Magnus Sesodia$^{1}$ \quad
  Juil Sock$^{2}$ \quad
  Philip Torr$^{1}$ \quad
  Christian Schroeder de Witt$^{1}$\\
  $^{1}$University of Oxford \quad $^{2}$BBC R\&D
}
\else
\author{
Magnus Sesodia \quad
Philip Torr \quad
Christian Schroeder de Witt\\
University of Oxford
}
\fi

\begin{document}

\maketitle

\begin{abstract}
Solving algorithmic tasks with many sequential steps typically requires scaling up transformer depth or chain-of-thought length, incurring unfavorable parameter and inference-compute costs, respectively.
Looped transformers, which reuse one block of layers across many iterations, are poised to address this, since they scale effective depth without adding parameters or tokens.
Yet, it remains unclear what each iteration is actually computing, or whether the loop merely simulates a deeper untied stack.
Here, we study \emph{directed permutation traversal}, in which a model follows $H$ hops from a starting node and reports the node it reaches, so every intermediate hop has a known ground truth.
We find that a minimum effective depth is necessary to solve the task, beyond which the degree of weight tying becomes the determining factor.
For instance, a looped transformer with a four-layer block iterated eight times solves the task in every seed, while an identically trained untied transformer of the same effective depth never does, succeeding only with a lower learning rate or supervision of its intermediate layers.
Probing and intervening on the residual stream, we find that the looped model advances through the intermediate nodes gradually, forming a staircase that adapts to its iteration budget, whereas the untied model, at that lower rate, exposes them only in its final layers.
Taken together, our results recast looped transformers as iterative program learners, which allocate their iterations to successive steps of the algorithm and solve tasks that identically trained untied models of the same effective depth do not.
Code is available at \url{https://github.com/magnus-sesodia/insidetheloop}.
\end{abstract}

\begin{figure}[H]
\centering
\includegraphics[width=1.0\linewidth]{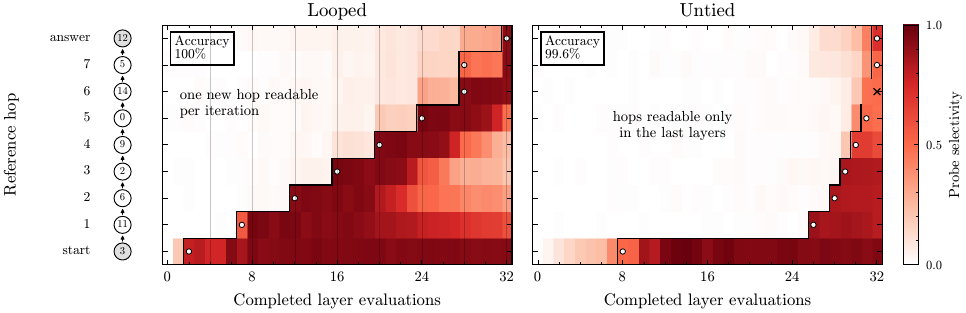}
\caption{
\textbf{Looped model forms a staircase.}
Linear-probe selectivity for each hop of the 8-hop \dePo\ task (illustrated at left), in a $4\times8$ looped model (left map) and an untied $32$-layer model (right map), both trained at $\eta=3\times10^{-4}$.
White dots mark where a hop first reaches selectivity $0.5$.
}
\label{fig:teaser}
\end{figure}

\section{Introduction}
\label{sec:intro}

Looped transformers have become an active front in latent reasoning thanks to their ability to scale computation without adding parameters or tokens.
Whereas an untied transformer applies each of its layers to an input just once ($T=1$), a looped model applies a shared block of layers $T \geq 2$ times by feeding the hidden state at the end of the block back to its start, multiplying that block's depth by $T$.
While the idea of a depth-recurrent transformer is not new \citep{dehghani_universal_2019, lan_albert_2020}, it had been sitting largely dormant until recent strong performance at multi-billion-parameter scale revived it \citep{geiping_scaling_2025a, zhu_scaling_2025_ouro}, triggering a proliferation of work spanning new variants \citep{bae_mor_2025, jeddi_loopformer_2026}, scaling laws \citep{prairie_parcae_2026, schwethelm_isodepth_2026} and theory \citep{merrill_logdepth_2025, xu_formal_2026}.

The most obvious benefit of looping is \emph{parameter efficiency}, since a block reused $T$ times occupies a $T$-th of the storage of an untied stack of the same depth.
A second advantage is \emph{higher reasoning bandwidth}, since a looped model keeps its intermediate computation in continuous hidden states, never compressing it into discrete chain-of-thought tokens \citep{xu_formal_2026, hao_coconut_2024}.
But the strength of looped models we investigate here is their purported \emph{inductive bias toward iterative computation} \citep{saunshi_latent_2025, giannou_looped_2023}.
Logically, weight tying should favor learning a repeated function, predisposing looped models to multi-hop reasoning and step-by-step algorithms.
Yet it is unclear whether a looped model actually runs such an algorithm, advancing the solution by a fixed number of steps per iteration, or merely simulates a deeper untied stack with fewer parameters.
The distinction matters for model design, since if the loop runs an algorithm, its iteration budget sets how many steps each iteration must take.
So far, mechanistic studies of hidden-state trajectories have not settled this question \citep{lu_latent_2025, chen_loop_bridge_2026, blayney_mechanistic_2026, curth_relational_2026}, in part because natural text offers no ground truth for intermediate states.

We work in a controlled setting using \emph{directed permutation traversal} (\dePo), a task from the physics-of-LMs lineage \citep{allenzhu_polm_4p1_2025, ye_implicit_reasoning_2025}.
Here, the model is presented with the edges of a cycle of $N$ nodes in shuffled order and reports the node reached after $H$ hops from a given start node (Figure~\ref{fig:methods-overview}).
This gives us a clear difficulty knob in the number of hops required $H$ and, crucially, every intermediate hop $h$ has a known node.
Thus, we can ask not just whether the final answer is correct but where in the network each intermediate node becomes decodable, although none is ever supervised.
This separates the two explanations, as an iterative algorithm predicts that each iteration adds a fixed number of hops, while a simulated deep stack need not tie nodes to iterations.

We begin by asking: \emph{do looped models outperform untied models at matched effective depth?}
If they do, a natural explanation is that weight tying steers training toward an iterative algorithm, in which one update is reused at every iteration.
Such an algorithm should leave evidence in the model's intermediate states, motivating the follow-up question: \emph{does each loop iteration advance the computation by a fixed number of hops?}
We answer it using linear probes on the hidden state after every layer, the model's own logit lens, and activation patching.

\paragraph{Contributions.}
\begin{enumerate}
  \item \textbf{Looped models are better learners.}
    We show that, trained identically at matched effective depth, looped models solve the task where untied models do not, even with eight times the parameters.
    Untied models need a lower learning rate or extra supervision to solve it (\S\ref{sec:h1}).
  \item \textbf{Looped models build a staircase (and use it).}
    We find that looped models advance the computation gradually with each iteration (Figure~\ref{fig:teaser}), at a rate set by the iteration budget, without any intermediate supervision.
    The untied models that solve the task show no such staircase unless their intermediate layers are supervised.
    Activation patching shows that the model causally uses the staircase for its answer (\S\ref{sec:h2}).
\end{enumerate}

\section{Related Work}
\label{sec:related}

\paragraph{Looped architectures.}
Universal Transformers \citep{dehghani_universal_2019} were the first to introduce the concept of a depth-recurrent transformer equipped with an adaptive halting mechanism \citep{graves_adaptive_2016}.
Beyond parameter efficiency \citep{lan_albert_2020}, looping can turn even a shallow transformer into a programmable computer \citep{giannou_looped_2023} and may bias learning toward iterative algorithms \citep{yang_looped_2024a, fan_looped_2024}.
More recently, at-scale work has established looped models as competitive on downstream reasoning benchmarks, looping either a middle block between an untied prelude and coda \citep{geiping_scaling_2025a, prairie_parcae_2026, mcleish_retrofitted_2025}, termed middle-cycle sharing by \citet{bae_mor_2025}, or the full stack \citep{zhu_scaling_2025_ouro, bae_rrt_2025, gao_urm_2025}.
Other variants add elastic depth \citep{jeddi_loopformer_2026}, multi-resolution recursion \citep{yu_spiralformer_2026}, per-token adaptive depth \citep{bae_mor_2025, luo_itt_2025}, iteration-aware conditioning \citep{xu_tmlt_2025}, hierarchical recurrence \citep{wang_hierarchical_2025} and hyper-connected residual streams \citep{zeitoun_hyperloop_2026}, skip or repeat pretrained layers \citep{li_skip_loop_2025} or stabilize many-step recurrence \citep{chen_thinking_deeper_2026}.
Iso-depth scaling laws also measure what one extra iteration is worth in added layers \citep{schwethelm_isodepth_2026}.
Ours is the simplest case, sharing every layer with an iteration encoding and no prelude, coda or input injection, to isolate the loop.

\paragraph{Theory and dynamics.}
\citet{saunshi_latent_2025} show that an $L$-layer transformer looped $T$ times nearly matches an $LT$-layer model on reasoning and prove that each loop can simulate one step of chain of thought.
\citet{merrill_logdepth_2025} prove that depth logarithmic in input length solves graph connectivity, which fixed depth cannot under standard conjectures.
Looping also parallelizes better than chain of thought \citep{xu_formal_2026}, yet expressivity alone cannot tell whether a trained loop runs an iterative algorithm or simulates the deeper stack.
A weight-tied layer converges to a fixed point \citep{bai_deq_2019}, and trained looped models settle into cyclic fixed points that repeat the stages of feedforward inference \citep{blayney_mechanistic_2026}, refine in small steps beneath a larger drift \citep{pappone_twoscale_2025} and resemble depth-grown models \citep{kapl_growing_looping_2026}.
Lens decoding of Huginn-3.5B, however, finds little evidence of a per-iteration chain \citep{lu_latent_2025}.
Untied transformers instead spread step-by-step search across layers and struggle to learn it \citep{brinkmann_multistep_2024, biran_hopping_2024, saparov_search_2025}, the contrast our comparison at matched effective depth isolates.

\paragraph{Closest work.}
\citet{kohli_loop_think_generalize_2026} study depth-recurrent transformers on multi-relation knowledge-graph walks, and we differ in task (single-permutation \dePo{} with every edge in the prompt), training and per-iteration probing.
Concurrently, \citet{zhang_recurrence_algorithm_2026} find on group word problems that weight tying installs a serial computation frontier resolving a fixed number of positions per loop, at a rate set by the training budget, which corroborates the per-loop rate we read from the hidden states (\S\ref{sec:h2}).

\section{Methods}
\label{sec:methods}

\subsection{Task: directed permutation traversal}
\label{sec:methods-task}

\begin{figure}[t]
\centering
\includegraphics[width=\linewidth]{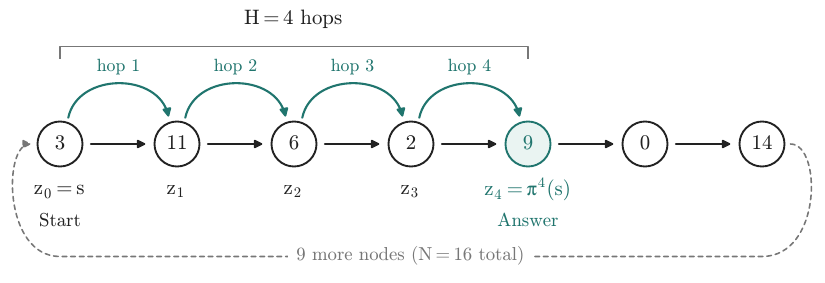}
\caption{\textbf{Directed permutation traversal.}
The model reads the edges of a cycle in shuffled order and must follow them for $H$ hops from the start node $s$.
In this example $s=3$ and $H=4$, so the walk visits the reference states $z_1=11$, $z_2=6$ and $z_3=2$ and ends at the answer $z_H=9$.
Straight arrows are edges and curved arrows are hops, and the dashed segment stands for the nine nodes not drawn, which complete the $N=16$ cycle.
}
\label{fig:methods-overview}
\end{figure}

We focus on directed permutation traversal (\dePo, Figure~\ref{fig:methods-overview}), a controlled depth-bound algorithmic task in the physics-of-LMs lineage \citep{allenzhu_polm_4p1_2025}.
A random permutation $\pi$ over $N=16$ nodes is sampled uniformly from the single $N$-cycles and presented to the model as an explicit edge list in shuffled order, followed by a query with a start node $s$ and a hop count $H$ whose answer is $\pi^H(s)$, the node $H$ steps along the cycle.
Since every hop depends on the result of the previous one, any correct walk visits the intermediate \emph{reference states} $z_h=\pi^h(s)$ for $h\in\{1,\ldots,H-1\}$, whose known ground truth lets us ask where in the network each one becomes decodable (\S\ref{sec:methods-probes}).
We train and evaluate on $H\in\{1,\ldots,15\}$, and five queries share the edge list of each prompt.

Because the cycle closes, $H$ forward hops reach the same node as $N-H$ backward hops, so a query needs at most $\min(H,N-H)\leq8$ sequential steps and we analyse internal states at $H\leq8$.
We check that the $2\times16$ model walks forward where the backward walk is shorter (App.~\ref{app:controls}, Fig.~\ref{fig:directional}) and repeat the capability comparison at $N=32$, where any $H\leq15$ takes $H$ steps (\S\ref{sec:h1-corrob}).

\subsection{Looped and untied models}
\label{sec:methods-arch}

A looped transformer reuses the same block of $L$ decoder layers across $T$ \emph{iterations}, and we write $L\times T$ for such a model.
With $g_{\phi}$ a decoder layer with parameters $\phi$, $f_{\theta}$ an $L$-layer block with parameters $\theta$ and $x^{(0)}$ the embedded input tokens, the two families compute
\begin{align}
\text{looped:}\quad x^{(t)} &= f_{\theta}\bigl(x^{(t-1)}+\tau_t\bigr), && t=1,\ldots,T, \label{eq:looped}\\
\text{untied:}\quad h^{(\ell)} &= g_{\phi_\ell}\bigl(h^{(\ell-1)}\bigr), && \ell=1,\ldots,LT, \label{eq:untied}
\end{align}
where $h^{(0)}=x^{(0)}$, $\tau_t$ is a learnable iteration embedding \citep{dehghani_universal_2019,xu_tmlt_2025} (ablated in App.~\ref{app:controls}), and an unembedding head is then applied to $x^{(T)}$ or $h^{(LT)}$.
The index $t$ counts iterations of one shared block, and $\ell$ counts distinct layers.
Per forward pass both perform $\Deff=LT$ layer evaluations, the \emph{effective depth}, but the untied model stacks $LT$ distinct layers applied once, so we call it the untied $LT$-layer model and place it at $T=1$ in Figure~\ref{fig:capability}.
That figure covers effective depths from $2$ to $32$, with looped blocks of $L\in\{2,4,8,16\}$ layers iterated up to $T=16$ times and untied models of $2$ to $32$ layers.
Both families are decoder-only with width $d=256$, $8$ heads, pre-norm RMSNorm, SwiGLU MLPs and rotary position embeddings (RoPE), and neither uses adaptive halting \citep{graves_adaptive_2016,banino_pondernet_2021}, input re-injection or intermediate readouts, so the two families differ mainly in how many distinct layers fill the same number of layer evaluations.
Besides weight tying, the looped model alone adds iteration embeddings and zero-initializes the output projections of its attention and MLP sublayers \citep{kohli_loop_think_generalize_2026}, and at matched $\Deff$ it has about $T$ times fewer parameters, e.g., $3.22$M for the $4\times8$ model against $25.72$M for the untied $32$-layer model.

\subsection{Training and evaluation}
\label{sec:methods-train}

All models are trained from scratch for $100$k steps with AdamW at peak learning rate $\eta=10^{-3}$ unless stated otherwise, with five seeds per configuration (App.~\ref{app:hyperparams}).
The loss is next-token cross-entropy on every non-pad token of the prompt, read out from the final layer only, so no intermediate hop is ever supervised.
A hop curriculum holds $H=1$ for the first $7$k steps and then widens the range of sampled hop lengths linearly to $H\leq15$ by step $30$k, after which every batch covers all $15$ hop lengths uniformly.
Section~\ref{sec:h1} also trains the looped $4\times8$ and the untied models at $\eta=3\times10^{-4}$ and $10^{-4}$, and, in one variant, adds the same next-token loss at the outputs of layers $8$, $16$ and $24$ of the untied $32$-layer model, read out through the shared final RMSNorm and unembedding with a total weight of $0.3$.
At $N=32$ we train three seeds per model at $\eta=3\times10^{-4}$ with a $384$-token context.
On trained models we score the first query of each prompt, and an answer counts as correct only if every digit and the delimiter that follows it are correct under teacher forcing, so always writing the same node scores $1/16$.
The evaluations every $4$k steps during training instead score the digits of the last query, whose earlier queries share $H$ and can repeat its start node, so a model that copies earlier answers scores about $0.36$ on this measure (App.~\ref{app:hyperparams}).
A seed \emph{solves} the task when its final training-time evaluation reaches accuracy $0.8$ on every hop length from $1$ to $15$, and its solved-chain depth $\Hsol$ is the largest $H$ such that every hop length up to $H$ reaches $0.8$ on that evaluation.

\subsection{Probes, logit lens and activation patching}
\label{sec:methods-probes}

\begin{figure}[t]
  \centering
  \includegraphics[width=\linewidth]{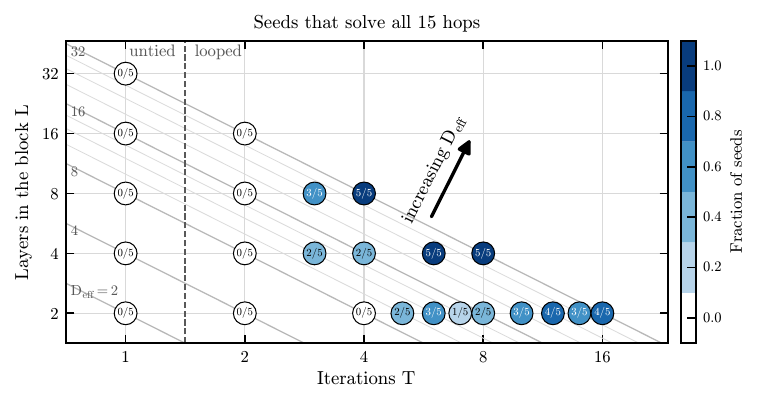}
    \caption{
    \textbf{Capability at learning rate \boldmath$\eta = 10^{-3}$.}
    Each node is a model with $L$ layers in its block, iterated $T$ times, trained with five seeds.
    Its label and color give the fraction of seeds that solve all $15$ hops, i.e., reach accuracy $0.8$ on every hop length from $1$ to $15$.
    Grey diagonals mark constant effective depth $\Deff{=}LT$.
    The $T{=}1$ column, left of the dashed line, holds the untied models.}
  \label{fig:capability}
\end{figure}

\paragraph{Linear probes.}
For every depth $\ell$ and hop $h$, we train a linear classifier to decode the reference state $z_h$ from the residual stream at the position just before the answer, i.e., the space token after the question mark, which predicts the answer's first digit.
We read the state after every layer, so layer $\ell$ of an $L\times T$ model ends iteration $\ell/L$, e.g., layer $8$ of the $4\times8$ model ends iteration $2$.
The probes use the five queries of $5{,}000$ fresh prompts, split by prompt into $4{,}000$ for fitting and $1{,}000$ held out, so no permutation appears on both sides.
As the headline quantity we use probe \emph{selectivity} \citep{hewitt_liang_2019}, $S_{\ell,h}=\mathrm{Acc}_{\mathrm{real}}(\ell,h)-\mathrm{Acc}_{\mathrm{shuffled}}(\ell,h)$, the gain in held-out accuracy over a control probe fitted to shuffled labels, which approaches one for a linearly readable node and zero otherwise.
A hop counts as \emph{decodable} wherever $S_{\ell,h}\geq0.5$, and the first depth at which it does is its \emph{onset}.
The untied maps at $\eta=10^{-3}$ (Figure~\ref{fig:organisation}b,d) and the forward-walk control use an earlier split by query row (App.~\ref{app:probes}).

\paragraph{Logit lens and activation patching.}
To see what the model would write after $t$ iterations, we stop the loop there and apply its own final RMSNorm and unembedding head \citep{nostalgebraist_logit_lens_2020}, scoring each node in full as the probability of all its digits and the delimiter, so the lens has no learned readout and reproduces the model's answers at $t=T$ (App.~\ref{app:probes}).
To check that the model relies on these states, we replace the receiver's states at the readout position and first answer digit after iteration $t$ with those of a donor prompt that shares the permutation but starts from a different node, and let the remaining iterations finish (Figure~\ref{fig:use}c).
An untrained model serves as the control, and App.~\ref{app:controls} repeats the patching with a donor that has a different permutation and the same start node.

\section{Looped models are better learners}
\label{sec:h1}

So why does looping work?
The simplest hypothesis is that looping buys effective depth, and
that this depth \emph{alone} explains the additional
capability.
If so, an $L$-layer model looped $T$ times and an untied
$LT$-layer model should perform about the same and, if
anything, the untied model should hold a slight advantage,
since it has $T$ times the parameters and so can express a much larger function class.
We put this to the test by jointly varying $L$ and $T$ at
several fixed values of $\Deff$, following the setup of
\citet{saunshi_latent_2025}.

\paragraph{Effective depth is necessary but not sufficient}
\label{sec:h1-matched}
We find, emphatically, that effective depth \emph{alone} cannot predict model capability.
This is clear when comparing models along any diagonal of
constant $\Deff$ in Figure~\ref{fig:capability}.
For example, at $\Deff = 32$ (the rightmost diagonal), the $8\times4$, $4\times8$ and $2\times16$ looped models solve the task in at least four of five seeds, while the untied $32$-layer model and the $16\times2$ model, which loops only twice, never do.
Thus, at fixed effective depth, how the depth is split between block size and iterations decides whether the task is learned.
The $\Deff = 16$ diagonal shows the same pattern more weakly, with $4\times4$ and $2\times8$ solving the task in two of five seeds while $8\times2$ and the untied $16\times1$ never do.
Indeed, at learning rate $10^{-3}$, no untied model solves the task at any depth from $2$ to $32$ layers.
Effective depth still matters, since no configuration at or
below $\Deff = 8$ solves the task.
This gap does not reflect undertraining, since the untied $32$-layer model's training loss changes by less than $0.03$ between the end of the hop curriculum at $30$k steps and the end of training at $100$k (App.~\ref{app:undertraining}).

\paragraph{Learning rate.}
\label{sec:h1-rate}
Every run in Figure~\ref{fig:capability} uses $\eta = 10^{-3}$, the looped model's learning rate.
But this may unfairly penalize the untied model, which may suit a different rate.
Slowing down to $\eta = 3{\times}10^{-4}$, we find the untied $32$-layer model now solves the task in four of five seeds, and even the untied $16$-layer model solves it in one seed (Table~\ref{tab:capability-rates}).
The $4\times8$ looped model still solves in every seed, and reaches the solution sooner on average ($45$k training steps against $55$k).
Slowing down further to $\eta = 10^{-4}$, both models solve in only one of five seeds.
Thus, the looped model learns the task over a wider range of learning rates, extending to faster rates at which the untied model fails.

\paragraph{Intermediate supervision.}
\label{sec:h1-aux}
We find the untied model can also solve the task when we additionally supervise its intermediate layers.
Specifically, we revert to $\eta = 10^{-3}$ and add a next-token loss at layers $8$, $16$ and $24$ of the untied $32$-layer model, each read out through the shared unembedding, keeping everything else fixed.
In this setup, the untied model solves the task in two of five seeds (Table~\ref{tab:capability-rates}, Fig.~\ref{fig:every-seed}d).

\paragraph{Longer chains.}
\label{sec:h1-corrob}
On a cycle of $N = 16$ nodes, any hop beyond $H = 8$ can also be reached by walking $N - H$ steps backward, so solving all $15$ hops takes at most $8$ sequential steps (\S\ref{sec:methods-task}).
To test longer chains, we repeat the comparison at $N = 32$, where reaching hop $H \le 15$ takes $H$ steps, using $\eta = 3{\times}10^{-4}$ (the rate at which both families learn the $N = 16$ task).
The untied $32$-layer model never gets past the first hop in any of its three seeds, while the looped $4\times8$ model, with an eighth of the parameters, solves the first six hops in one of its three seeds (Table~\ref{tab:capability-rates}, Fig.~\ref{fig:n32}).

\begin{table}[t]
  \centering
  \caption{\textbf{Looped $4\times8$ against untied $32$-layer.} The untied column holds the $32$-layer model
  unless the row names another depth. Seeds that solve all $15$ hops at
  $N{=}16$, and solved-chain depth $\Hsol\!@0.8$ for each seed at $N{=}32$.}
  \label{tab:capability-rates}
  \small
  \begin{tabular}{llcc}
    \toprule
    Task & $\eta$ & Looped $4\times8$ & Untied $32$-layer \\
    \midrule
    $N{=}16$ & $10^{-3}$ & $5/5$ & $0/5$ \\
    $N{=}16$ & $3{\times}10^{-4}$ & $5/5$ & $4/5$ \\
    $N{=}16$ & $10^{-4}$ & $1/5$ & $1/5$ \\
    $N{=}16$, untied $16$ layers & $3{\times}10^{-4}$ & -- & $1/5$ \\
    $N{=}16$, auxiliary loss & $10^{-3}$ & -- & $2/5$ \\
    \midrule
    $N{=}32$ & $3{\times}10^{-4}$ & $\Hsol = 6, 0, 1$ & $\Hsol = 1, 1, 1$ \\
    \bottomrule
  \end{tabular}
\end{table}

\section{Looped models build a staircase (and use it)}
\label{sec:h2}

Having established that untied models can also solve the task, albeit only with a lower learning rate or auxiliary supervision, we ask whether they solve it in the same way as looped models.
The nature of the task, with its precisely defined intermediate states, provides a unique opportunity to answer this directly, by tracing where in the network each (intermediate) node becomes decodable (\S\ref{sec:methods-probes}).

\begin{figure}[t]
  \centering
  \includegraphics[width=0.8\linewidth]{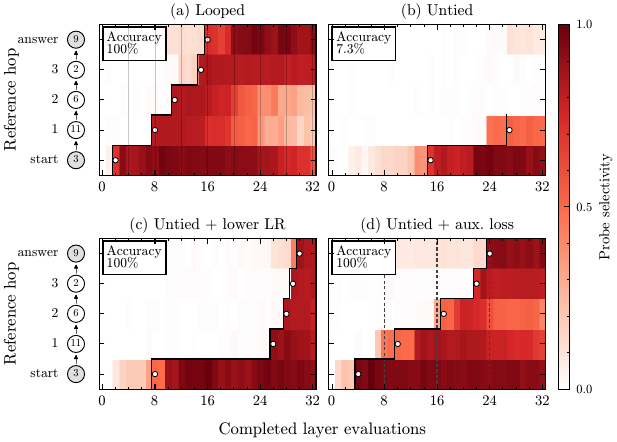}  %
    \caption{
      \textbf{Looped vs.\ untied solution structures.}
      Probe selectivity (\S\ref{sec:methods-probes}) by layer and hop at $H{=}4$:
      (a) the looped $4\times8$ model at $\eta = 10^{-3}$;
      (b) the untied $32$-layer model at the same rate;
      (c) the untied model at the lower rate $\eta = 3{\times}10^{-4}$;
      (d) the untied model at $\eta = 10^{-3}$ with an auxiliary next-token loss on the outputs of layers $8$, $16$ and $24$ (dashed lines).
      White dots, joined by black lines, mark the first layer at which each hop reaches selectivity $0.5$, and faint grey lines in (a) separate iterations.
      Onsets for the other seeds are shown in Figure~\ref{fig:every-seed}.
    }
  \label{fig:organisation}
\end{figure}

\paragraph{Solution structures.}
\label{sec:h2-design}
We examine how each model steps through the $H=4$ solution (if at all) in Figure~\ref{fig:organisation}.
The looped $4\times8$ model reveals successive intermediate nodes a few layers apart, gradually reaching the final answer node by layer $16$ (a).
The untied $32$-layer model trained at the same learning rate, however, does not solve the task, and beyond the start node at most one hop ever becomes decodable (b).
At the lower learning rate that lets the untied model solve the task (\S\ref{sec:h1-rate}), no node beyond the start becomes decodable for the first $25$ layers and then, in a sudden flurry, all three intermediate nodes and the answer become decodable between layer $26$ and $30$ (c).
Finally, with the auxiliary loss of \S\ref{sec:h1-aux}, the untied model builds a staircase resembling that of the looped model, but with non-uniform step sizes (d).
The same holds at $H=8$ (Figure~\ref{fig:teaser}) and across seeds (App.~\ref{app:probes}).

\begin{figure}[t]
  \centering
  \includegraphics[width=0.92\linewidth]{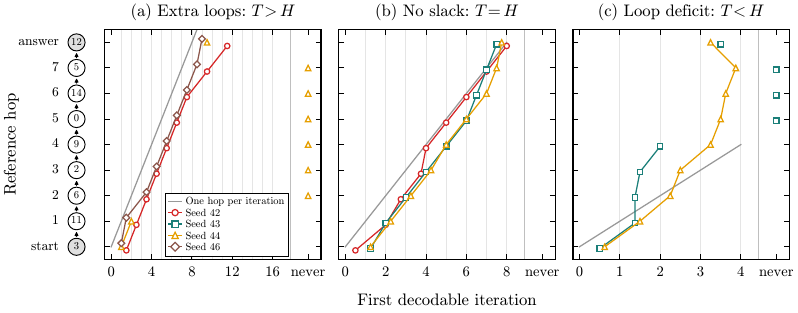}
  \caption{\textbf{How the loop spends its iterations.} The first iteration at which each reference node becomes
  decodable (selectivity $0.5$) at $H{=}8$, for looped models with more iterations than hops ($2\times16$, a), as many
  ($4\times8$, b) and fewer ($8\times4$, c). Fractional positions give the layer within the iteration, light vertical
  lines separate iterations. Markers in the never column are nodes that never become decodable. Each model was
  trained with five seeds, and a seed is drawn if its answer accuracy is at least $0.95$ (three seeds in a and b, two
  in c). All models are trained at $\eta = 10^{-3}$.}
  \label{fig:dynamics}
\end{figure}

\paragraph{Hopping within an iteration budget.}
\label{sec:h2-onset}
It is often purported that a looped model consumes one iteration per well-defined step of an algorithm \citep{giannou_looped_2023, saunshi_latent_2025, kohli_loop_think_generalize_2026}.
Here, we put this to the test by checking whether hop $h$ indeed becomes decodable at iteration $h$, in models whose iteration budget $T$ exceeds, equals, or falls short of the $H=8$ required hops of the task (Figure~\ref{fig:dynamics}).

With more iterations than hops ($T>H$), the $2\times16$ model advances about one hop per iteration in two of three seeds and finishes by iteration $12$ of $16$, and no further hop becomes decodable in the spare iterations (a).
The third seed actually reaches the answer by iteration $10$, although hops $2$ to $7$ never become decodable (App.~\ref{app:controls}).
With as many iterations as hops ($T=H$), the $4\times8$ model roughly follows the identity line in all three seeds (b).
However, it needs about two iterations to make the first hop, and later makes two hops  within a single iteration, so the walk still ends by iteration $8$.
With fewer iterations than hops ($T<H$), the $8\times4$ model makes several hops within single iterations (!) to reach the answer within its fourth and final iteration (c).
At $H=12$, where four backward steps would reach the answer, the $2\times16$ model still makes every forward hop in turn (App.~\ref{app:controls}, Fig.~\ref{fig:directional}).
Thus, in most seeds, a looped model takes about one hop per iteration when its budget allows, and packs several hops into an iteration when it does not.

\begin{figure}[t]
  \centering
  \includegraphics[width=0.85\linewidth]{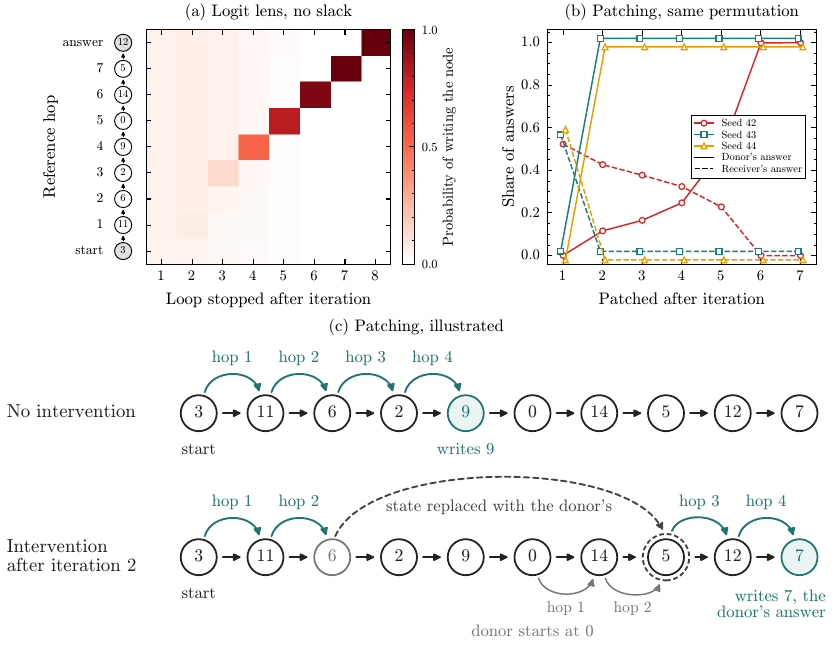}
  \caption{\textbf{Logit lens and causal interventions.}
  (a) Logit lens: probability that the looped $4\times8$ model ($H{=}8$, one seed, $\eta=10^{-3}$) writes each reference node when its loop is stopped after a given iteration and its state read through the final RMSNorm and unembedding head.
  (b) Share of answers that are the donor's (solid) or the receiver's (dashed) when the receiver's states at the readout position and first answer digit are replaced by the donor's after a given iteration ($4\times8$, $H{=}6$, three seeds, coinciding series offset). Donor and receiver share the permutation and differ in the start node.
  (c) Schematic of (b) at $H{=}4$: the receiver walks from $3$ and writes $9$, but with its state after iteration $2$ replaced by that of a donor starting at $0$, it finishes the donor's walk and writes $7$.}
  \label{fig:use}
\end{figure}

\paragraph{Logit lens and causal interventions.}
\label{sec:h2-use}
While probes reveal information that a model can represent linearly, they do not guarantee that the model actually uses it.
To test this, we first apply the model's own output head after each iteration and ask which node it would have written had that been the final iteration, i.e., the logit lens \citep{nostalgebraist_logit_lens_2020}.
In the $4\times8$ model, the head writes the node of hop $t$ after iteration $t$, with probability rising from $0.53$ at hop $4$ to $1.00$ at hop $7$ (Figure~\ref{fig:use}a).
This mirrors the probe staircase of Figure~\ref{fig:dynamics}, with one interesting difference.
The head never puts more than $0.10$ of its probability on nodes the walk has already passed, so the area under the staircase is white, whereas a probe still decodes each passed node at least half its peak selectivity for about two more iterations (App.~\ref{app:probes}).
Thus, a passed node fades from the residual stream gradually while the head stops writing it at once, which suggests that the head reads only the current node from a stream that still holds earlier ones.

We also perform causal interventions (Figure~\ref{fig:use}c), running the model on a second, donor prompt that asks about the \emph{same} permutation from a \emph{different} start node.
Specifically, we copy the donor's states at the readout position and first answer digit into the original run after a given iteration and let the remaining iterations finish (\S\ref{sec:methods-probes}).
If these states are meaningfully used, then the model should finish the donor's walk and write the donor's answer.
We find that the share of donor answers reaches $1.00$ once the patch comes after the second iteration in two of three seeds, and after the sixth in the third (Figure~\ref{fig:use}b), while an untrained model writes neither answer (App.~\ref{app:controls}).

Thus, the model's own output head can read the staircase \emph{and} the model completes whichever walk its state holds.
One caveat is that the logit lens tracks every hop in only one seed per iteration budget, and a state copied from a prompt with a different permutation yields the node composed from both walks in only one of five seeds, and in no seed of the $2\times16$ model (App.~\ref{app:controls}).

\section{Discussion and limitations}
\label{sec:discussion}
\label{sec:limitations}

\paragraph{Pretrained looped models.}
So why did \citet{lu_latent_2025} only find limited evidence of a per-iteration chain in Huginn-3.5B \citep{geiping_scaling_2025a}?
Part of the answer may be the readout, since \citeauthor{lu_latent_2025} decode the state with the logit lens and Huginn's coda but train no probes, and in our models the lens writes the current node at each iteration in only one seed per iteration budget, whereas linear probes decode every hop in most seeds at $H=4$ (App.~\ref{app:probes}).
The rest may lie in training, since Huginn samples a new number of iterations at each training step and backpropagates through only the last eight, whereas our models are each trained at one budget, to which the number of hops per iteration adapts (Figure~\ref{fig:dynamics}).
Our task also has a correct node at every step and trains every hop length, two conditions generic pretraining lacks, so we do not claim the staircase forms without them.

\paragraph{Learnability.}
That the untied $32$-layer model solves the task at a slower learning rate or with a next-token loss on its intermediate layers is evidence that it is capable of representing a solution (Table~\ref{tab:capability-rates}).
Supervising its intermediate layers also gives it a staircase (Figure~\ref{fig:organisation}d), and we leave to future work why tied training reaches the staircase while the untied stack needs intermediate supervision.

\paragraph{Design.}
For looped model practitioners, our results at a fixed learning rate show that no model looped fewer than three times solved the task (Figure~\ref{fig:capability}), but a model's budget need not cover every step of the task, since the $8\times4$ model solves it in every seed, and its probed seeds pack several hops into each iteration (Figure~\ref{fig:dynamics}c).
A model with spare iterations could exit once its walk ends \citep{graves_adaptive_2016, zhu_scaling_2025_ouro}, but whether running a model that packs hops for more iterations than it was trained with, as \citet{geiping_scaling_2025a} do for Huginn, extends its walk is untested here.

\paragraph{Limitations.}
First, we test a single synthetic task, \dePo, chosen because a per-iteration probe needs a known correct answer at every intermediate step, and although we corroborate the behavioral result at $N{=}32$ (\S\ref{sec:h1-corrob}) and concurrent work reports a per-loop rate set by the training budget on group word problems \citep{zhang_recurrence_algorithm_2026}, whether the staircase survives on tasks that mix retrieval with computation, or on natural language, is open.
Second, our models have at most ${\sim}26$\,M parameters and loop every layer, so our results concern inductive bias at small scale, and the at-scale regime where Huginn and Ouro \citep{zhu_scaling_2025_ouro} report their largest gains, with a prelude and coda around the loop \citep{geiping_scaling_2025a}, remains untested.
Third, the grid in Figure~\ref{fig:capability} uses a single learning rate, only its $4\times8$ and untied $32$-layer models were trained at three rates (Table~\ref{tab:capability-rates}), and we probe the internal states of only four of its configurations.

\section{Conclusion}
\label{sec:conclusion}

We are now in a position to answer the two questions of \S\ref{sec:intro}.
First, under identical training at matched effective depth, looped models solve the \dePo\ task while untied models, with eight times the parameters, do not, and succeed only with a lower learning rate or supervision of their intermediate layers (\S\ref{sec:h1}).
Second, in most seeds, looped models take about one hop per iteration when their budget allows and pack several hops into an iteration when it does not.
Linear probes decode these intermediate states and activation patching shows that the model uses them (\S\ref{sec:h2}).
Thus, on this task, where every intermediate step has a known answer and the loop's computation is therefore legible, looped transformers appear to be iterative program learners, which execute an algorithm across their iterations without any supervision of its intermediate steps, making the iteration budget a design parameter for tasks with many sequential steps.

\begin{ack}
We thank \ifarxiv\else Dr Juil Sock and \fi Dr Bruno Andreis for insightful discussions.
M.S.\ is supported by the Engineering and Physical Sciences Research Council (EPSRC) through an Industrial Doctoral Landscape Award (IDLA), grant UKRI2002.
The authors also gratefully acknowledge the matching support and collaboration provided by BBC Research \& Development.
P.T.\ is supported by the UKRI Turing AI Fellowship EP/W002981/1 and the Schmidt Sciences AI2050 Senior Fellowship.
This publication was supported by the Department for Science, Innovation and Technology and the Royal Academy of Engineering under the Research Fellowships scheme.
The authors declare no competing interests.
\end{ack}

\bibliographystyle{plainnat}
\bibliography{main}

\clearpage
\appendix
\renewcommand{\thefigure}{A\arabic{figure}}
\renewcommand{\thetable}{A\arabic{table}}
\renewcommand{\theHfigure}{A\arabic{figure}}  %
\renewcommand{\theHtable}{A\arabic{table}}
\setcounter{figure}{0}
\setcounter{table}{0}

\section{Training details}
\label{app:hyperparams}

\paragraph{Architecture.} Both families use width $d=256$, $8$ heads of dimension $32$, pre-norm RMSNorm ($\epsilon=10^{-5}$), SwiGLU MLPs of hidden width $704$ and rotary position embeddings. The vocabulary has $18$ tokens, ten digits and eight syntax and special tokens, so nodes $10$ to $15$ take two tokens. Looped models add the iteration embedding $\tau_t$ before each iteration and zero-initialize the attention and MLP output projections of their block \citep{kohli_loop_think_generalize_2026}, while untied models keep the default initialization. No model has input re-injection, a halting head, a prelude or a coda.

\paragraph{Optimization.} AdamW with $\beta=(0.9, 0.95)$, $\epsilon=10^{-8}$, weight decay $10^{-2}$ and gradient clipping at norm $1$. The learning rate warms up linearly over $1{,}000$ steps to its peak $\eta$ and then follows a cosine decay to $0.1\eta$. Every run trains for $100$k steps and is evaluated at its final checkpoint. Each step draws $120$ prompts spread evenly over the hop lengths the curriculum allows at that step (\S\ref{sec:methods-train}), left-padded to $200$ tokens. The five queries of a prompt share $H$ and draw their start nodes independently. At $N=32$ the peak rate is $3\times10^{-4}$, the warm-up $2{,}000$ steps and the context $384$ tokens, with three seeds per model and four for the untied $4$-layer control.

\paragraph{Evaluation during training.} Every $4$k steps we score $8{,}000$ fresh prompts from a held-out generator seed (about $533$ per hop length) on the digits of the last query's answer, under teacher forcing. Because the queries of a prompt share $H$ and can repeat a start node, a model that copies earlier answers of the prompt scores about $0.36$ on this measure, well below the $0.8$ that defines a solve.

\paragraph{Seeds and hardware.} Seeds are $42$ to $46$, and the untied $4$-layer control at $N{=}32$ also uses seed $47$. The seed sets the initialization, the batch order and the training data, and every cell shares one evaluation seed. Runs were trained on NVIDIA H200 and A40 GPUs. Training is deterministic on a given GPU type, but the same seed on the other type diverges within the first few hundred steps and, for some looped seeds, ends at a different solved-chain depth, so we read each cell's count as a sample of its seeds.

\begin{figure}[!htbp]
  \centering
  \includegraphics[width=0.6\linewidth]{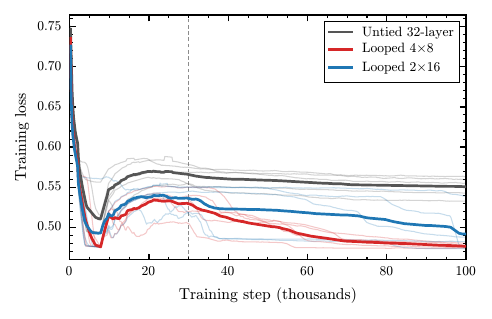}
  \caption{\textbf{Training loss at matched effective depth.} Training loss of the untied $32$-layer model and the looped $4\times8$ and $2\times16$ models at $\eta = 10^{-3}$, averaged over a trailing window of $2$k steps, for each of five seeds (thin) and their mean (thick). The dashed line marks the end of the hop curriculum at $30$k steps, before which the loss rises as longer chains enter training.}
  \label{fig:undertraining}
\end{figure}

\paragraph{Compute parity and undertraining.}
\label{app:undertraining}
At matched $\Deff$, looped and untied models perform the same number of layer evaluations per step and share the step budget, schedule, batch size and hop curriculum, so their FLOPs per step differ only by the addition of $\tau_t$. On the same H200 hardware, the looped $4\times8$, looped $2\times16$ and untied $32$-layer models each train in about $329$ minutes. At $\eta=10^{-3}$ the untied $32$-layer model's training loss falls by only $0.005$ to $0.023$ between the end of the hop curriculum at $30$k steps and the end of training, while it falls by $0.03$ to $0.07$ in every looped seed that solves the task (Fig.~\ref{fig:undertraining}).

\section{Probing details}
\label{app:probes}

\begin{figure}[!htbp]
  \centering
  \includegraphics[width=\linewidth]{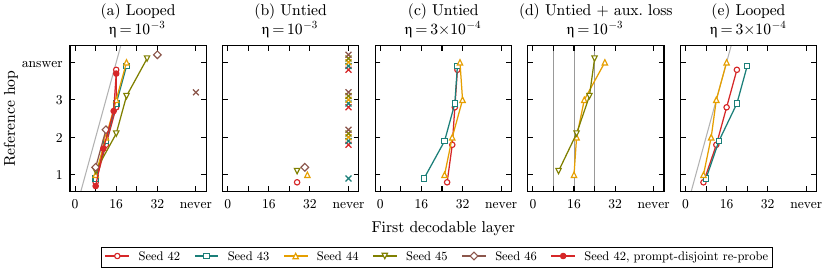}
  \caption{\textbf{Where each hop first becomes decodable, every seed.} First layer at which each hop of the
  $H{=}4$ chain reaches probe selectivity $0.5$, one colour and marker per seed. Crosses in the never column mark hops
  that never do, and seeds are offset vertically by up to $0.2$ hops so that coinciding points stay visible.
  (a) Looped $4\times8$ at $\eta = 10^{-3}$, seeds $42$ to $46$ (row split, read at the end of each iteration, so
  onsets fall on multiples of $4$), and the prompt-disjoint re-probe of seed $42$ (filled). (b) Untied $32$-layer
  model at $\eta = 10^{-3}$ (row split). (c) Untied at $\eta = 3{\times}10^{-4}$ (prompt-disjoint). (d) Untied with
  the auxiliary loss on layers $8$, $16$ and $24$ (grey vertical lines), $\eta = 10^{-3}$, the two seeds that solve
  the task (row split). (e) Looped $4\times8$ at $\eta = 3{\times}10^{-4}$ (prompt-disjoint). The grey diagonal in
  (a) and (e) is one hop per iteration.}
  \label{fig:every-seed}
  \label{fig:deep32-layerprobe}
\end{figure}

\paragraph{Probe protocols.} Every probe reads the residual stream at the query position of $5{,}000$ evaluation prompts with five queries each ($25{,}000$ query rows), and one linear probe $\mathrm{Linear}(d, N)$ is fitted per state level and hop, for $30$ epochs with AdamW (learning rate $10^{-3}$, weight decay $10^{-4}$, batch size $256$). A $32$-layer computation has $33$ state levels, the embedding and the state after each layer. The prompt-disjoint protocol of \S\ref{sec:methods-probes} splits the prompts $4{,}000/1{,}000$ and fits the shuffled-label control on the training prompts only. An earlier protocol splits the $25{,}000$ query rows at random $80{:}20$, so queries of one prompt can fall on both sides, and reads looped models only at the end of each iteration ($9$ levels for $4\times8$, $17$ for $2\times16$). Figs.~\ref{fig:teaser} and \ref{fig:dynamics}, Fig.~\ref{fig:organisation}(a, c) and Fig.~\ref{fig:every-seed}(c, e) use prompt-disjoint probes, as does the seed-$42$ re-probe in Fig.~\ref{fig:every-seed}(a). Fig.~\ref{fig:organisation}(b, d), the other runs of Fig.~\ref{fig:every-seed}(a, b, d), Fig.~\ref{fig:directional} and Fig.~\ref{fig:anomalous}(a) use the row split. Every accuracy reported with the probes is full-answer accuracy on the first query of each of the $5{,}000$ prompts (\S\ref{sec:methods-train}).

\paragraph{Every seed.} Fig.~\ref{fig:every-seed} and Table~\ref{tab:every-seed} give the $H{=}4$ onsets of every probed seed of the four conditions in Fig.~\ref{fig:organisation}, and of the looped $4\times8$ model at the lower rate. The auxiliary-loss condition was probed in the two of its five seeds that solve the task. In every looped seed the decodable hops appear in hop order, and the answer first becomes decodable in iteration $4$ to $8$ at $\eta = 10^{-3}$ and in iteration $4$ to $6$ at $3{\times}10^{-4}$. Hop $3$ of seed $46$ at $10^{-3}$, which solves the task on the training-time evaluation but answers $65\%$ of first queries, never reaches selectivity $0.5$. In the untied model at $10^{-3}$, where no seed solves the task (full-answer accuracy $0.07$ to $0.17$), at most hop $1$ becomes decodable, at layer $27$ or later. At $3{\times}10^{-4}$ every hop of the untied model becomes decodable from layer $25$ on, apart from hop $1$ of seed $43$ at layer $17$, and with the auxiliary loss from layer $10$ or $16$ on.

\begin{table}[!htbp]
  \centering
  \small
  \caption{\textbf{First decodable layer and accuracy, every seed.} First layer at which each hop at $H{=}4$ reaches
  selectivity $0.5$ (``--'' never) and full-answer accuracy on the first query, for every seed of
  Fig.~\ref{fig:every-seed}. Looped onsets in the row split are resolved per iteration.}
  \label{tab:every-seed}
  \begin{tabular}{llcccccc}
    \toprule
    Model, $\eta$ & Probes & Seed & Hop 1 & Hop 2 & Hop 3 & Answer & Accuracy \\
    \midrule
    Looped $4\times8$, $10^{-3}$ & row split & 42 & 8 & 12 & 16 & 16 & 1.000 \\
     & row split & 43 & 8 & 12 & 16 & 20 & 1.000 \\
     & row split & 44 & 8 & 12 & 16 & 20 & 1.000 \\
     & row split & 45 & 8 & 16 & 20 & 28 & 1.000 \\
     & row split & 46 & 8 & 12 & -- & 32 & 0.650 \\
     & prompt-disjoint & 42 & 8 & 11 & 15 & 16 & 1.000 \\
    \addlinespace
    Untied, $10^{-3}$ & row split & 42 & 27 & -- & -- & -- & 0.073 \\
     & row split & 43 & -- & -- & -- & -- & 0.074 \\
     & row split & 44 & 31 & -- & -- & -- & 0.083 \\
     & row split & 45 & 27 & -- & -- & -- & 0.087 \\
     & row split & 46 & 30 & -- & -- & -- & 0.165 \\
    \addlinespace
    Untied, $3{\times}10^{-4}$ & prompt-disjoint & 42 & 26 & 28 & 29 & 30 & 1.000 \\
     & prompt-disjoint & 43 & 17 & 25 & 29 & 30 & 1.000 \\
     & prompt-disjoint & 44 & 25 & 28 & 32 & 31 & 0.982 \\
    \addlinespace
    Untied + aux., $10^{-3}$ & row split & 44 & 16 & 17 & 20 & 28 & 0.855 \\
     & row split & 45 & 10 & 17 & 22 & 24 & 1.000 \\
    \addlinespace
    Looped $4\times8$, $3{\times}10^{-4}$ & prompt-disjoint & 42 & 7 & 12 & 16 & 20 & 1.000 \\
     & prompt-disjoint & 43 & 8 & 13 & 20 & 24 & 1.000 \\
     & prompt-disjoint & 44 & 7 & 10 & 12 & 16 & 1.000 \\
    \bottomrule
  \end{tabular}
\end{table}

\paragraph{Retention.} After its peak, a hop's selectivity stays at or above half that peak for a number of further iterations, which we call its retention. In the looped $4\times8$ model at $H{=}8$, averaged over hops $1$ to $7$ (the answer has no iterations left), retention is $2.2$ iterations for the three seeds at $\eta = 3{\times}10^{-4}$ (Fig.~\ref{fig:teaser}) and $2.6$ for the prompt-disjoint re-probe of seed $42$ at $10^{-3}$, both read per layer and converted to iterations, and $1.7$ for seeds $42$ to $45$ at $10^{-3}$ under the row split, read per iteration.

\paragraph{Sanity controls.} Running the same probes (row split, $H{=}4$) on an untrained looped $4\times8$ model gives accuracies within $0.008$ of chance in every cell, so the probe signal comes from the trained weights. Probes retrained on shuffled labels \citep{hewitt_liang_2019} never exceed accuracy $0.19$ in any cell of either protocol (chance $1/16$). Stopped after the last iteration, the logit lens of \S\ref{sec:methods-probes} reproduces the model's own full-answer accuracy exactly in all $14$ probed runs of the models in Fig.~\ref{fig:dynamics} at $H{=}8$ and all $10$ at $H{=}4$, and an untrained model never writes a node under the lens (probability at most $0.002$).

\section{Additional controls}
\label{app:controls}

\begin{figure}[!htbp]
  \centering
  \includegraphics[width=0.6\linewidth]{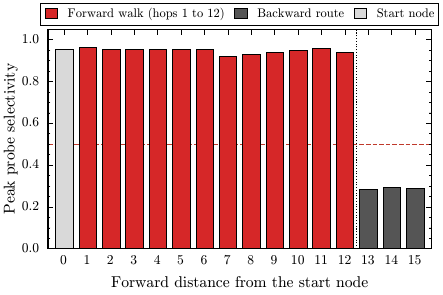}
  \caption{\textbf{Forward-walk nodes are decodable.} Peak probe selectivity for each node $\pi^{d}(s)$ at forward distance $d$ from the start node $s$, in the looped $2\times16$ model at $H{=}12$ ($\eta = 10^{-3}$, row split). The dotted line separates the forward walk from the three nodes of the four-step backward route, and the dashed line is the decodability threshold $0.5$.}
  \label{fig:directional}
\end{figure}

\paragraph{Forward walk.} At $H{=}12$ the answer is also four backward steps from the start. Probes on the looped $2\times16$ model (seed $42$, $\eta = 10^{-3}$, row split, $15{,}000$ queries) decode each of the twelve forward-walk nodes $\pi^{d}(s)$, $d=1,\ldots,12$, at peak selectivity $0.92$ to $0.96$, with the peak iteration rising from $4$ at the second node to $16$ at the twelfth (the first node peaks at iteration $7$), while the three nodes of the backward route peak at $0.29$ (Fig.~\ref{fig:directional}). The same pattern holds at $H{=}10$.

\begin{figure}[!htbp]
  \centering
  \includegraphics[width=0.6\linewidth]{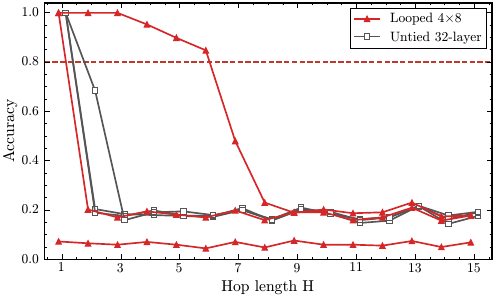}
  \caption{\textbf{Longer chains at $N{=}32$.} Per-hop accuracy of the evaluation during training (App.~\ref{app:hyperparams}) at the final step, for each of three seeds of the looped $4\times8$ model and the untied $32$-layer model at $\eta = 3{\times}10^{-4}$. Points are offset slightly in $H$ so that overlapping seeds stay visible. The dashed line marks the $0.8$ threshold that defines a solved hop.}
  \label{fig:n32}
\end{figure}

\paragraph{$N{=}32$.} Every run trains for $100$k steps at $\eta = 3{\times}10^{-4}$ with a $384$-token context (Fig.~\ref{fig:n32}). Solved-chain depths at the final step are $6$, $0$ and $1$ for the looped $4\times8$ model (seeds $42$ to $44$) and $1$ in every seed of the untied $32$-layer model. An untied $4$-layer model, with about as many parameters as the looped model, reaches at most $\Hsol{=}1$ in four seeds. Since $H \leq 15 < N/2$, no query admits a backward walk shorter than the forward one.

\begin{figure}[!htbp]
  \centering
  \includegraphics[width=\linewidth]{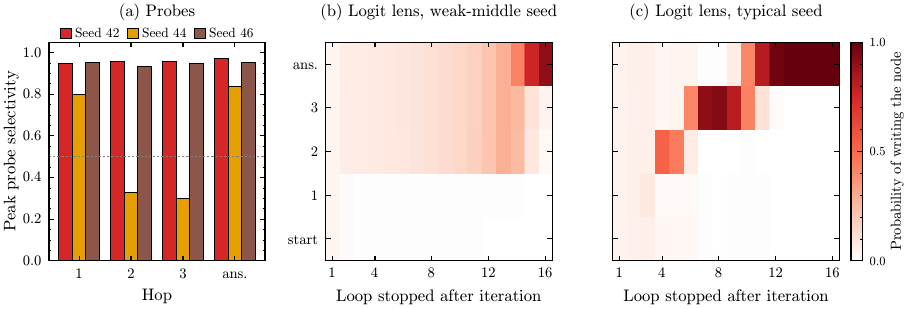}
  \caption{\textbf{A looped seed with weak middle hops.} (a) Peak probe selectivity per hop at $H{=}4$ for seeds $42$, $44$ and $46$ of the looped $2\times16$ model at $\eta = 10^{-3}$ (row split, read after each iteration), with the dotted line at the decodability threshold $0.5$. (b, c) Probability that seed $44$ (b) and seed $42$ (c) write each reference node when the loop is stopped after a given iteration, read through the model's own head (\S\ref{sec:methods-probes}), averaged over $5{,}000$ prompts.}
  \label{fig:anomalous}
\end{figure}

\paragraph{A seed with weak middle hops.} Seed $44$ of the looped $2\times16$ model at $\eta = 10^{-3}$, the third seed of Fig.~\ref{fig:dynamics}(a), answers the $H{=}4$ task (full-answer accuracy $0.959$) while hops $2$ and $3$ stay below selectivity $0.5$ (Fig.~\ref{fig:anomalous}a). Its peak selectivity for hops $1$ to $3$ and the answer is $0.80$, $0.33$, $0.30$ and $0.84$, against $0.95$, $0.96$, $0.96$ and $0.97$ for seed $42$. Under the logit lens, seed $44$ never writes hop $2$ or $3$ alone, and until iteration $13$ it spreads its probability over hops $2$ and $3$ and the answer ($0.28$ each at iteration $13$), and it writes the answer with probability $0.91$ only after the last iteration, while seed $42$ writes hop $2$, hop $3$ and the answer in turn, at peak probabilities $0.53$, $0.93$ and $1.00$ (Fig.~\ref{fig:anomalous}b, c). At $H{=}8$ the same seed leaves hops $2$ to $7$ undecodable while answering $99\%$ of first queries.

\paragraph{Logit lens, every seed.} We apply the full-node lens at $H{=}8$ to the checkpoints of Fig.~\ref{fig:dynamics}, on $5{,}000$ prompts each. In one seed per iteration budget the head writes the node the walk has reached, as the $4\times8$ seed of Fig.~\ref{fig:use}(a) writes hop $t$ after iteration $t$ ($0.53$ to $1.00$ for hops $4$ to $7$), seed $42$ of the $2\times16$ model writes hop $6$ after iteration $8$ ($0.77$) and hop $7$ after iterations $10$ to $12$ (up to $0.98$), and seed $43$ of the $8\times4$ model writes hop $4$ after iteration $2$ ($0.46$) and hop $6$ after iteration $3$ ($0.41$). In the other seeds the head writes only the answer, gives the nodes the walk has passed almost no probability, and reaches lens accuracy $0.9$ only after the last iteration when $T \le H$, and after iteration $10$ (seed $46$) or $13$ (seed $44$) in the $2\times16$ model.

\begin{figure}[!htbp]
  \centering
  \includegraphics[width=0.75\linewidth]{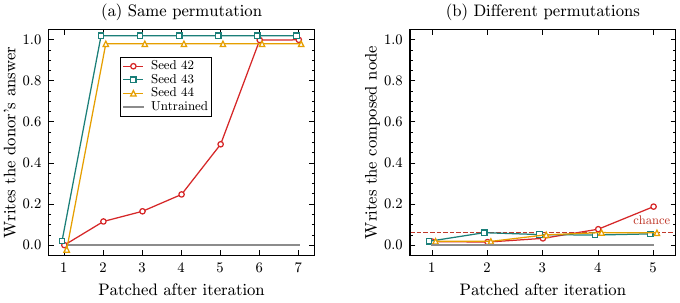}
  \caption{\textbf{Activation patching, three seeds.} Looped $4\times8$ model at $H{=}6$, seeds $42$ to $44$, with the states at the readout position and first answer digit replaced after a given iteration (\S\ref{sec:methods-probes}). (a) Same permutation: fraction of the $2{,}000$ answers that are the donor's. (b) Different permutations: fraction of answers that are the node composed from both walks, among the $1{,}721$ to $1{,}780$ answers where it differs from both prompts' answers. A patch after iteration $6$ or later makes the composed node the donor's answer, so (b) ends at iteration $5$. The dashed line is chance ($1/16$), grey lines are an untrained model, and coinciding seeds are offset slightly.}
  \label{fig:patching-every-seed}
\end{figure}

\paragraph{Activation patching, every seed.} We score all five queries of $400$ prompt pairs, with earlier answers teacher forced, for the looped $4\times8$ model at $H{=}6$, and of $300$ pairs for the $2\times16$ model at $H{=}8$. With a shared permutation, every answer is the donor's once the patch comes after the second iteration in seeds $43$ and $44$, and after the sixth in seed $42$ (Fig.~\ref{fig:patching-every-seed}a). Of the two $4\times8$ seeds the figure omits, seed $45$ gives $0.97$ after the second iteration and at least $0.998$ thereafter, and seed $46$, which answers $76\%$ of queries at $H{=}6$ without patching, gives $0.46$ to $0.70$. With different permutations, the node composed from both walks rises above chance only in seed $42$, to $0.19$ after iteration $5$, and stays at or below $0.06$ in the four other $4\times8$ seeds (Fig.~\ref{fig:patching-every-seed}b) and in every seed of the $2\times16$ model. An untrained model writes neither answer in either test. Patching only the readout position moves the first digit, but the second digit of a two-digit answer then becomes the donor's in under a tenth of cases, so we also patch the first answer digit.

\paragraph{Iteration-embedding ablation.} Retraining the looped $4\times8$ model at $\eta = 10^{-3}$ without the iteration embedding $\tau_t$ \citep{xu_tmlt_2025} leaves three of its four seeds solving the task (five of five with it), so the task is learned without it. In a prompt-disjoint re-probe at $H{=}4$ and $H{=}8$, every hop of the three solving seeds becomes decodable, in hop order.

\end{document}